\pdfoutput=1
\documentclass[11pt]{article}
\usepackage{acl}
\usepackage{times}
\usepackage{latexsym}
\usepackage{enumitem}
\usepackage[T1]{fontenc}
\usepackage[utf8]{inputenc}
\usepackage{microtype}
\usepackage{graphicx}
\usepackage{csquotes}
\usepackage{hyperref}

\newcommand{\bilal}[1]{\textcolor{black}{#1}}

\usepackage{multicol}
\usepackage{multirow}
\usepackage{dcolumn}
\newcolumntype{d}[1]{D{.}{.}{#1}}

\title{Question Generation for Reading Comprehension Assessment by Modeling How and What to Ask}

\author{
Bilal Ghanem$^{1}$, Lauren Lutz Coleman$^{2}$, Julia Rivard Dexter$^{2}$, \\ \textbf{Spencer McIntosh von der Ohe$^{1}$, Alona Fyshe$^{1}$} \\
  {\normalsize $^{1}$University of Alberta, Canada}\\
  {\normalsize $^{2}$EyeRead, Canada} \\
  \small{\{bghanem,vonderoh,alona\}@ualberta.ca}\\ 
  \small{\{lauren,julia\}@eyeread.co}
}

\begin{document}
\maketitle

\begin{abstract}
Reading is integral to everyday life, and yet learning to read is a struggle for many young learners. During lessons, teachers can use comprehension questions to increase engagement, test reading skills, and improve retention. Historically such questions were written by skilled teachers, but recently language models have been used to generate comprehension questions.  
%The process of asking questions is important for both humans and machines, as it helps knowledge acquisition and can test comprehension. One application is tests of reading comprehension, where a human or a machine asks a set of questions to readers to help them learn and engage with the text. 
However, many existing Question Generation (QG) systems focus on generating \emph{literal} questions from the text, and have no way to control the type of the generated question. In this paper, we study QG for reading comprehension where \emph{inferential} questions are critical and extractive techniques cannot be used. We propose a two-step model (HTA-WTA) that takes advantage of previous datasets, and can generate questions for a specific targeted comprehension skill. We propose a new reading comprehension dataset that contains questions annotated with story-based reading comprehension skills (SBRCS), allowing for a more complete reader assessment. Across several experiments, our results show that HTA-WTA outperforms multiple strong baselines on this new dataset. We show that the HTA-WTA model tests for strong SCRS by asking deep inferential questions.
\end{abstract}

\section{Introduction}
Reading is an invaluable skill, and is core to communicating in our digital age.  Reading also supports other forms of development; when children read, it sharpens their memory, and improves social skills~\cite{halliday1973explorations,mason2017reading}. Yet, statistics show that one out of five children in the U.S. face learning difficulties~\cite{shaywitz2005overcoming}, especially in reading~\cite{cornoldi2013reading}. The coronavirus pandemic beginning in 2020 had a huge impact on the early reading skills of many children, and threatens to leave a lasting impact on a whole generation of young readers~\cite{gupta2020impacts}.

The pandemic forced many children to learn online, putting in sharp relief the need for effective online education platforms. In particular, reading games have become popular, and can help fill the gap when teachers cannot read in person with students. These platforms present students with short passages and associated comprehension questions.
These questions are key to assessing a reader's comprehension of a passage, and can also enhance learning~\cite{chua2017mediated}. But, writing diverse and engaging comprehension questions is a non-trivial task.

Teachers need to generate new comprehension questions whenever they incorporate new text into a curriculum.  New text helps to keep material fresh and topical, and can allow teachers to customize lessons to the interests of a particular student cohort. After finding such custom reading material, teachers must write new comprehension questions to evaluate several reading aspects of comprehension (e.g. understanding complex words, recalling events, etc.). 

Thus, to improve the educational process, and lighten the load on teachers, we need tools to automate Question Generation (QG): the task of writing questions for a given passage. 
%One distinctive side of QG systems is the type of questions that they produce, where 
Generated questions can be either inferential or literal (extractive) questions. Literal questions can be answered using only information stated in the text, whereas inferential questions require additional information or reasoning. Previous works focused on this aspect of the questions in reading comprehension and discarded the comprehension skills (e.g. close reading, predicting, figurative language, etc.)~\cite{murakhovs2021mixqg}.

We take inspiration from continual learning~\cite{parisi2019continual}, which orders a set of learning tasks to improve model performance. We begin by training a model on the general task of QG (How to ask: HTA), and follow with our task of interest: generating a targeted question of a particular type (What to ask: WTA).

This paper focuses on the generation of questions for story-based reading comprehension skills (SBRCS), which are varied and cover many aspects of reading comprehension. We create a QG dataset for SBRCS\footnote{We are working with our industrial partner to publish the dataset once it is completed as we are still working on incorporating more SBRCS. The dataset will be published only for research purposes.}. Although our aim in creating this dataset is to enrich educational applications, this dataset can be considered as a source for general QG and question answering (QA) systems in NLP. 

Our focus here is to build a question generator without answer supervision as the case in a real-life application, where a story only will be given as input. This is a challenging task, as many different questions can be generated from a story when there is no answer supervision. QG with answer supervision is another prevalent research line in the literature~\cite{zhao2018paragraph, ma2020improving, wang2020answer, chen2021answer}.

The contributions in this work are as follows: 
\begin{itemize}
\itemsep0em 
    \item \noindent We build a novel QG dataset for SBRCS. The dataset contains advanced reading comprehension skills extracted from stories.
    
    \item \noindent We propose a two-steps method to generate skill-related questions from a given story. The method takes advantage of previous datasets to improve generalizability, and then, teaches a model how to ask predefined styles of questions.
    
    \item \noindent We demonstrate the efficiency of the proposed method after extensive experiments, and we investigate its performance in a few-shot learning setting.
\end{itemize}

The rest of the paper is structured as follows. In the next section, we present an overview of the literature work. In Section \ref{sec:data}, we describe how we built our dataset. Section \ref{sec:methodology} describes the proposed methodology. The experimental setting is presented in Section \ref{sec:experiments}. The results and the analysis are presented in Section \ref{sec:results:analysis}.  Finally, we draw some conclusions and possible future work for this study.

% \vspace{0.5em}
\section{Related Works}
\label{sec:related:works}
% \noindent \textbf{Question Generation}
QG has progressed rapidly due to new datasets and model improvements. Many different QG models have been proposed, starting for simple vanilla Sequence to Sequence Neural Networks models (seq2seq)~\cite{du2017learning,zhou2017neural,yuan2017machine} to the more recent transformer-based models~\cite{dong2019unified,chan2019recurrent,varanasi2020copybert,narayan2020qurious,bao2020unilmv2}. Some QG systems use manual linguistic features in their models~\cite{harrison2018neural,khullar2018automatic,liu2019learning,dhole2020syn}, some consider how to select question-worthy content~\cite{du2017identifying,li2019improving,scialom2019self,liu2020asking}, and some systems explicitly model question types~\cite{duan2017question,sun2018answer,kang2019let,zhou2019question}. The last group focused only on generating questions that start with specific interrogative words (what, how, etc.).

QG has been used to solve many real-life problems. For example, QG in conversational dialogue~\cite{gu2021chaincqg,shen-etal-2021-gtm,liu2021learning} where models were taught to ask a series of coherent questions grounded in a QA style, QG based on visual input~\cite{mostafazadeh2016generating,shin2018customized,shukla2019should}, and QG for deep questions such as mathematical, curiosity-driven, clinical, and examination-type questions~\cite{liyanage2019multi,scialom2020ask,cliniqg4qa2020,jia2021eqg}.

% QG Datasets subsection if there is space.

% \vspace{0.5em}
\section{Data}
\label{sec:data}
Despite the recent efforts for building reading comprehension QA datasets, to the best of our knowledge, none of the available datasets explored SBRCS. Questions in previous datasets ask only either inferential or literal questions from a given passage/story. \citet{rogers2020getting}, developed questions with general reasoning types based on text from news and blogs \bilal{(e.g. Quora)}. We believe that those texts sources are not rich enough to examine reasoning skills. Advanced reasoning skills (e.g. Figurative Language) are usually used in \bilal{children's} stories to assess comprehension skills. \bilal{Additionally, we use a extensive set of reading comprehension skills that deeply evaluates the abilities of the readers (e.g. imagination skill by \textit{Visualizing})}.  In the following, we will show how we built our dataset. Table \ref{tab:data_stat} gives an overview of the dataset.

\subsection{Dataset Design}
\label{subsec:dataset_design}

\subsubsection{Stories Collection}
Our stories (passages) are multi-genre, self-contained narratives. This content variety leads annotators towards asking non-localized questions that test for more advanced reading comprehension skills. The stories are generated using several resources: 1. acquired from free public domain content (Gutenberg Project\footnote{\href{https://www.gutenberg.org/}{https://www.gutenberg.org/}}), 2. partnerships with a publishing house (Blue Moon Publishers\footnote{\href{https://bluemoonpublishers.com/}{https://bluemoonpublishers.com/}}) and an educational curriculum development foundation (The Reimagined Classroom\footnote{\href{https://www.reimaginedclassroom.com/}{https://www.reimaginedclassroom.com/}}), and 3. authored by two professional writers, 
(the majority of the stories are from this last category). To provide good lexical coverage and diverse stories, we choose to write and collect stories that come from a varied set of genres (e.g. science, social studies, fantasy, fairy tale, historical fiction, horror, mystery, adventure, etc.). In total, we collect 726 multi-domain stories. The stories' lengths range from a single sentence to 113 sentences. 

\begin{table*}%[h]
\small
\centering
\begin{tabular}{l|d{4.1}|d{3.1}|d{3.1}|d{3.1}|d{3.1}|d{3.1}|d{3.1}|d{3.1}|d{3.1}}
\hline
& \textbf{BSE} & \textbf{CT} & \textbf{CR} & \textbf{FL} & \textbf{I} & \textbf{P} & \textbf{S} & \textbf{V} & \textbf{VO} \\
\hline
\# Stories               & 269.0  & 280.0  & 448.0  & 219.0  & 449.0  & 152.0  & 360.0 & 153.0  & 403.0  \\
\# Question–answer pairs & 390.0  & 415.0  & 719.0  & 292.0  & 695.0  & 162.0  & 560.0 & 163.0  & 604.0  \\
Avg. \#tok. in stories   & 168.98 & 189.62 & 133.44 & 137.86 & 133.63 & 145.09 & 192.8 & 118.61 & 143.21 \\
Max. \#tok. in stories   & 1159.0 & 1159.0 & 1159.0 & 935.0  & 1159.0 & 1132.0 & 1132.0 & 935.0  & 1040.0 \\
Avg. \#tok. in questions & 9.14   & 11.82  & 11.12  & 16.38  & 13.21  & 12.92  & 9.88  & 12.98  & 15.96  \\
Max. \#tok. in questions & 24.0   & 58.0   & 55.0   & 70.0   & 52.0   & 76.0   & 43.0  & 39.0   & 49.0   \\
Avg. \#tok. in answers   & 4.17   & 3.81   & 4.49   & 4.7    & 6.16   & 6.48   & 5.91  & 5.10   & 3.46   \\
Max. \#tok. in answers   & 29.0   & 34.0   & 73.0   & 30.0   & 29.0   & 21.0   & 46.0  & 40.0   & 22.0   \\
\# Literal Questions     & 274.0  & 120.0  & 606.0  & 108.0  & 16.0   & 11.0   & 464.0 & 36.0   & 168.0  \\
\# Inferential Questions & 115.0  & 295.0  & 113.0  & 148.0  & 679.0  & 151.0  & 96.0  & 127.0  & 436.0   \\
\hline
\end{tabular}
\caption{\label{tab:data_stat} Collected dataset's statistics. There are 726 stories, which can have questions from multiple skill types (described in Section \ref{subsec:dataset_design}).}
\end{table*}

\subsubsection{Questions and Comprehension Skills}
Previous comprehension question datasets focused on either inferential or literal questions. Although these questions assess comprehension skills, they do not provide fine-grained evaluation of the reader comprehension. Thus, to build a more comprehensive list of question types, we started by reviewing curriculum documents available from Columbia University Teacher's College Readers\footnote{\href{https://www.tc.columbia.edu/curriculum-and-teaching/literacy-specialist/the-reading--writing-project/}{https://www.tc.columbia.edu/curriculum-and-teaching/literacy-specialist/the-reading--writing-project/}} and Writers Workshop Program\footnote{\href{https://readingandwritingproject.org/}{https://readingandwritingproject.org/}}. Then, we compiled a list of SBRCS, which we then expanded to include additional skills based on school teachers' recommendations. In Section \ref{sec:appendix:skills:details}, we present further details for each skill type. \bilal{Also, in Appendix \ref{sec:appendix:taxonomy}, we give further details on the skills list and on the educational theory behind the skills taxonomy}. Our final list contains the following skills:

\begin{enumerate}
\itemsep0em
    \item \noindent \textbf{Basic Story Elements (BSE)}: Can the reader identify the story's main characters and setting? 
        \begin{displayquote}
            \centering
            \textit{\bilal{From the details in this passage, how many individuals were part of this investigation?}}
        \end{displayquote}
    
    \item \noindent \textbf{Character Traits (CT)}: Can the reader identify the traits attributable to certain characters in the story (e.g. character feelings, physical attributes)? 
        \begin{displayquote}
            \centering
            \textit{\bilal{How did the Rabbit feel in this passage?}}
        \end{displayquote}
    
    \item \noindent \textbf{Close Reading (CR)}: Can the reader extract the text span in a story where the author best describes or explains a key point?
        \begin{displayquote}
            \centering
            \textit{\bilal{How many people celebrated Karata's birth?}}
        \end{displayquote}
    
    \item \noindent \textbf{Figurative Language (FL)}: Is the reader able to recognize the implied meaning of a sentence?
        \begin{displayquote}
            \centering
            \textit{\bilal{Reread this sentence: ``His legs were pumping so fast that they felt like jelly.'' What did the author mean by this?}}
        \end{displayquote}

    \item \noindent \textbf{Inferring (I)}: Can the reader infer what happened in between scenes if the time in-between is not explicitly described?
        \begin{displayquote}
            \centering
            \textit{\bilal{Why do you think Minho opened the suitcase?}}
        \end{displayquote}
    
    \item \noindent \textbf{Predicting (P)}:
    Can the reader find textual clues and use them to guess what would happen next?
        \begin{displayquote}
            \centering
            \textit{\bilal{Do you think that the bear enrolled in classes and became a student?}}
        \end{displayquote}
    
    \item \noindent \textbf{Summarizing (S)}: Is the reader able to recognize the main literary elements of the characters, the events, the problem, and the solutions?
        \begin{displayquote}
            \centering
            \textit{\bilal{What is Bal doing?}}
        \end{displayquote}
    
    \item \noindent \textbf{Visualizing (V)}: Can the reader visualize scenes in her/his head to fully comprehend the story?
        \begin{displayquote}
            \centering
            \textit{\bilal{What is the author trying to describe by writing ``everything below became smaller and smaller''?}}
        \end{displayquote}
    
    \item \noindent \textbf{Vocabulary (VO)}: Can the reader identify the right meaning of a word within a context when the word has multiple possible definitions?
        \begin{displayquote}
            \centering
            \textit{\bilal{Which word in the passage is a synonym for ``stubborn''?}}
        \end{displayquote}
    
\end{enumerate}

With our list of SBRCS as a guide, we wrote question-answer pairs for each story. Given the difficulty of the task, we needed a large number of trained content writers to build the required questions. Each written question should fall into one of the mentioned skills. For that, a total of 25 professionals contributed to the writing process (18 teachers, 7 graduate students). \bilal{Each annotator was asked to write a question per skill for a given story. Not every skill is applicable to every story, so some skills were discarded for some stories.} We chose not to use crowdworkers (e.g. Amazon Mechanical Turk) to ensure high-quality and educationally-appropriate questions. To verify the quality of the generated content, a second team member reviews each question-answer pair before adding them to the dataset. \bilal{If the second team member found issues, a discussion took place. In the cases that the team members could not reach an agreement, a third team member is brought in to resolve the disagreement}. In addition to annotating questions with a skills label, our content writers annotate each question as either \textit{Literal} or \textit{Inferential} question types. This information is important to measure the comprehension performance of the reader on each question type. Overall, we generate 4K question-answer pairs, with an average of 5.5 pairs per story. \bilal{Note that we did not ask multiple annotators to write questions per story in order to measure the annotators' agreement. Different annotators often write the same question in different ways, or may choose a different question topic for a given skill, or even select a different skill. Thus, measuring inter-annotator agreement is not meaningful. Instead, we chose to ask one annotator to write questions and another to validate the questions grammatically and to check whether the question is correctly related to the chosen skill.}

% \vspace{0.5em}
\section{Methodology}
\label{sec:methodology}

Given the fact that including more data in a reading comprehension system is important for generalization~\cite{chung2018supervised,talmor-berant-2019-multiqa}, and given that our created dataset has the SBRCS which are missed in previous datasets, we propose a two-steps method to generate skill-related questions from a given story: HTA followed by WTA. HTA teaches the model the typical format for comprehension questions using large previously released datasets. \bilal{We use two well-known datasets, SQuAD~\cite{rajpurkar2016squad} and CosmosQA~ \cite{huang2019cosmos}. In Appendix \ref{sec:appendix:additional_data}, we add more details on both of these datasets.} These previous datasets are not annotated with the question types outlined in Section~\ref{subsec:dataset_design}, so the HTA phase allows us to take advantage of those datasets. WTA guides the model to generate questions to test the specific comprehension skills enumerated in Section \ref{subsec:dataset_design}. Thus, in HTA, we train (fine-tune) a model on large QG datasets, and then, we further train the model to teach the model what to ask (WTA). For the generation model, we use the pre-trained Text-to-Text Transfer Transformer T5~\cite{raffel2020exploring}, which closely follows the encoder-decoder architecture of the transformer model~\cite{Vaswani2017attention}. T5 is a SOTA model on multiple tasks, including QA.

\subsection{How to Ask (HTA)}
Previous works showed that incorporating more data when training a reading comprehension model improves performance and generalizability~\cite{chung2018supervised,talmor-berant-2019-multiqa}. However, we cannot incorporate previously released datasets with our new one, as they do not include compatible question skills information. However, they do contain many well-formed and topical questions.  Thus, we train a T5 model on SQuAD and CosmosQA datasets to teach the model \emph{how} to ask questions.  

Previous neural question generation models take the passage as input, along with the answer. However, encoders can pass all of the information in the input to the decoder, occasionally causing the generated question to contain the target answer. Since the majority of the questions in our created dataset are inferential questions, the answers are not explicitly given in the passages (unlike extractive datasets). Thus, we feed the stories to the encoder, but withhold the answers. Unlike previous systems, we then train the model to generate the questions and \emph{answers}. We propose this setting to generate fewer literal questions. During our experiments, we evaluated the effect of excluding the answers, and we found them useful to the system.

\begin{figure}%[H]
\centering
\includegraphics[width=7.7cm]{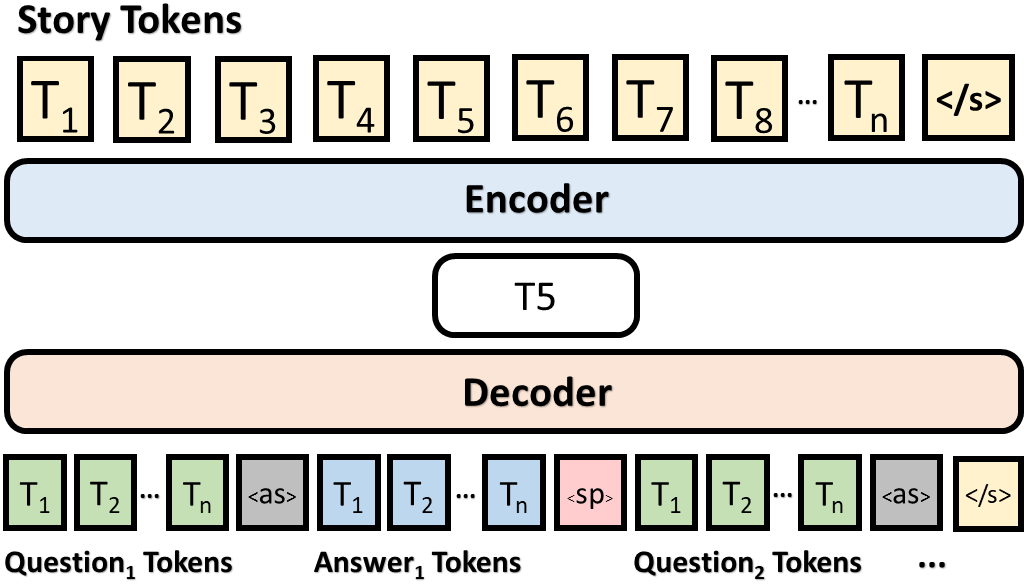}
\caption{Input and output format of the \textbf{How to Ask} (\textbf{HTA}) model.}
\label{fig:HTA_arch}
\end{figure}

In Figure \ref{fig:HTA_arch} we show the input-output format of the model. The encoder input is structured as {\small \textit{<STORY\_TEXT>}} {\small \textit{</s>}}, where {\small \textit{</s>}} is the end-of-sentence token. The decoder generates multiple question-answer pairs as {\small \textit{<QUESTION\_TOKENS>$_1$} \textit{<as>} \textit{<ANSWER\_TOKENS>$_1$} \textit{<sp>} ... \textit{<QUESTION\_TOKENS>$_n$} \textit{<as>} \textit{<ANSWER\_TOKENS>$_n$} \textit{</s>}}, where {\small \textit{<as>}} separates a question from its answer, and {\small \textit{<sp>}} separates a question-answer pair from another. The model can generate more than one question-answer pair. We prepare the data to include all of a passage's question-answer pairs in the decoder. Some passages include single question-answer pair, and some passages have up to fifteen pairs.

\subsection{What to Ask (WTA)}
QG models take a passage/story as input and generate a question. The type of generated question is not controlled and is left for the system to decide it. Thus, the generated question is usually an undesired question. Thus, in order to control the style of the generated question, the system needs an indication about the skill that the system is expected to generate a question for. \citet{liu2020asking} proposed a way to control the style of the generated questions (e.g. what, how, etc.). The authors built a rule-based information extractor to sample meaningful inputs from a given text, and then learn a joint distribution of <answer, clue, question style> before asking the GPT2 model~\cite{radford2019language} to generate questions. However, this distribution can only be learned using an extractive dataset (e.g. SQuAD); the model cannot learn to generate inferential questions. 

\begin{figure}%[H]
\centering
\includegraphics[width=7.7cm]{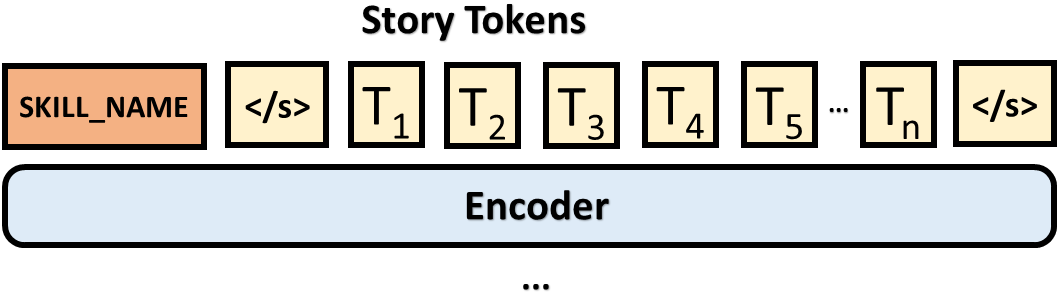}
\caption{Input format of the \textbf{What to Ask} (\textbf{WTA}) model. The output format is the same as in HTA model (see Figure \ref{fig:HTA_arch}).}
\label{fig:WTA_arch}
\end{figure}

To control the skill of the generated question, we use a specific prompt per skill, by defining a special token {\small \textit{<SKILL\_NAME>}} corresponding to the desired target skill, \bilal{using the collected dataset}. This helps us to control what to extract from the pretrained model. Thus, the encoder takes as input {\small \textit{<SKILL\_NAME>}} and {\small \textit{<STORY\_TEXT>}}, where {\small \textit{<SKILL\_NAME>}} indicates to the model for which skill the question should be generated (see Figure \ref{fig:WTA_arch}). The data format in the decoder is similar to the one in the HTA step, but here the model generates a single question-answer pair. As a result, the encoding of the {\small \textit{<STORY\_TEXT>}} will be based on the given {\small \textit{<SKILL\_NAME>}}. In this way, the model encodes the same story in a different representation when a different {\small \textit{<SKILL\_NAME>}} is given. A similar technique was used in the literature to include persona profiles in dialogue agents to produce more coherent and meaningful conversations~\cite{scialom2020toward}.

% \vspace{0.5em}
\section{Experiments}
\label{sec:experiments}

\subsection{Decoding Method}
Decoding strategies are crucial and directly impact output quality. In general, Beam Search~\cite{dept._2018} is the most common algorithm, in addition to some other sampling techniques such as Nucleus sampling (Top-p)~\cite{holtzman2019curious}. In Beam Search, the output of a model is found by maximizing the model probability. On the other hand, Nucleus sampling selects the smallest possible set of tokens whose cumulative probability exceeds the probability p. Experimentally, we found that using the top-p (p=0.9) algorithm yields the best results in terms of the used scoring metrics, thus we use it in all of our experiments.

\subsection{Evaluation Metrics}
QG often uses standard evaluation metrics from text summarization and machine translation (BLEU~\cite{papineni2002bleu}, ROUGE~\cite{lin2004rouge}, METEOR~\cite{banerjee2005meteor}, etc.). However, such metrics do not provide an accurate evaluation for QG task~\cite{novikova2017we}, especially when the input passage is long (and many acceptable questions that differ from the gold question can be generated). Thus, to alleviate shortcomings associated with n-gram based similarity metrics, we use BLEURT~\cite{sellam2020bleurt} (\textit{BLEURT-20}), which is state-of-the-art evaluation metric in WMT Metrics shared task. BLEURT is a BERT-based model that uses multi-task learning to evaluate a generated text by giving it a value mostly between 0.0 and 1.0. In our experiments, we consider BLEURT as the main metric for the evaluation. We also report standard MT metric BLEU (1-4 ngrams), and perform an additional manual evaluation. 

Manual evaluation is required in our collected dataset, because teachers wrote a single question per skill for a given story, where the model might generate other possible questions for the same skill.

\subsection{Implementation Details} 
We fine-tune a T5 model (\textit{t5-base} from HuggingFace library) using the Adam optimizer with a batch size of 8 and a learning rate of $1e-4$. We use a maximum sequence length of 512 for the encoder, and 128 for the decoder\footnote{We were restricted to this length due to memory shortage.}. We tested the T5-large model, but we did not notice any improvements considering BLEURT metric. We train all models for a maximum of ten epochs with an early stopping value of 1 (patience) based on the validation loss. We use a single NVIDIA TITAN RTX with 24G RAM. 

For HTA, we validate on a combined version of the validation sets from both datasets (SQuAD and CosmosQA). Regarding the collected dataset validation set, we use stratified sampling: we took a random 10\% of stories from each skill since the dataset is unbalanced. We apply the same strategy with the test set but with a value of 20\%.

\subsection{Baselines} 
To evaluate the performance of our
model, we use a set of models that showed state-of-the-art results on several datasets. We obtain the results of those models by running their published GitHub code on our collected dataset. For all of the following baselines, we use SQuAD, CosmosQA, and the collected dataset for training and we test on the test part of the collected dataset:
\begin{itemize}
\itemsep0em 
    \item \noindent Vanilla Seq2seq~\cite{sutskever2014sequence}:  a basic
encoder-decoder sequence learning system for machine translation. This model takes the story as input and generates a question.

    \item \noindent NQG-Seq~\cite{du2017learning}: another Seq2seq that implements an attention layer on top of a bidirectional-LSTM encoder. The authors use two encoders, one to encode the sentence that has the answer, and another to encode the whole document. The model then is trained to generate questions.
    
    \item \noindent NQG-Max~\cite{zhao2018paragraph}\footnote{We used the unofficial implementation in this GitHub repo: \href{https://github.com/seanie12/neural-question-generation}{https://github.com/seanie12/neural-question-generation}}: a QG system with a maxout pointer mechanism and gated self-attention LSTM-based encoder to address the challenges of processing long text input. This model takes a passage and an answer as input and generate a question. The answer must be a sub span of the passage. 
    
    \item \noindent CGC-QG~\cite{liu2019learning}: a Clue Guided Copy network for Question Generation, which is a sequence-to-sequence generative model with a copying mechanism that takes a passage and an answer (as a span in the text) and generate the question. The text representation in the encoder (GRU network) is represented using a variety of features such as GloVe vectors, POS information, answer position, clue word, etc.
    
    \item \noindent AnswerQuest~\cite{roemmele2021answerquest}: a pipeline model that uses as a first step a previous model~\cite{yang-etal-2019-end-end} to retrieve the relevant sentence that has the answer from a document. And then, the sentence is fed to a transformer-based sequence-to-sequence model that is enhanced with a copy mechanism. 
    
    \item \noindent One-Step: a baseline that uses T5 model trained with all data in one step instead of having separate HTA and WTA steps. Because there is only a single step, the skill name is not included in the encoder's input.
    
    \item \noindent T5-WTA: the WTA model trained using T5 model as a seed model. The HTA training step is not used here. We use this baseline to evaluate the effect of training WTA using HTA.
    
\end{itemize}

For all of the previous baselines that require the answer to be a sub-span in the passage, we use the semantic text similarity method that was proposed in~\cite{ghanem2019upv} to retrieve the most similar span in the passage. The method extracts several ngrams features from a claim and text spans, and then compute cosine similarity to get the most similar span. In this work, we replace the ngrams features of a text with embeddings extracted from RoBERTa model~\cite{yinhan2019roberta}. This process has been done on the inferential questions as their answers are not clearly given in the text.

% \vspace{0.5em}
\section{Results and Analysis}
\label{sec:results:analysis}
Table \ref{tab:results_main} presents the results of the proposed \textit{HTA-WTA} method with the baselines. We can see that out of the baselines, \textit{T5-WTA} performs best in terms of BLEURT score (32.96\%), followed by \textit{NQG-Max} with a value of 31.78\%. Given its high BLEURT score, it is surprising that \textit{T5-WTA} model has low BLEU-4. This implies that the generated questions use rich vocabulary, making them different from the gold in terms of overlapping ngrams, but semantically similar leading to higher BLEURT score. As shown in the table, \textit{HTA-WTA}'s BLEURT score outperforms all of the previous QG models by a noticeable margin, showing that including the skill name information plays an important role in generating the intended questions. Also, training on more QG datasets improves the performance. \bilal{We also noted that the \textit{CGC-QG} model achieves a higher BLEU-1 than our \textit{HTA-WTA} model. We argue that this is because the Clue Words Prediction Module learns important cues, increasing the uni-gram overlap with the gold references (BLEU-1).}

\begin{table*}%[h]
\small
\centering
\begin{tabular}{l|d{2.2}|d{2.2}|d{2.2}|d{2.2}|d{2.2}}
\hline
\textbf{Model} & \multicolumn{1}{|c|}{\textbf{BLEU-1}} & \multicolumn{1}{|c|}{\textbf{BLEU-2}} & \multicolumn{1}{|c|}{\textbf{BLEU-3}} & \multicolumn{1}{|c|}{\textbf{BLEU-4}} & \multicolumn{1}{|c|}{\textbf{BLEURT}} \\ \hline
Vanilla Seq2seq     & 17.16 & 7.78 & 4.28 & 2.37 & 08.42 \\
NQG-Seq             & 18.85 & 8.31 & 4.37 & 2.49 & 11.13 \\
NQG-Max             & 19.27 & 7.17 & 4.12 & 2.77 & 31.78 \\
CGC-QG & \textbf{23}.\textbf{93} & 12.01 & 7.82 & 5.68 & 29.28 \\
AnswerQuest         & 20.44 & 9.08 & 4.53 & 4.71 & 29.15 \\
One-Step            & 15.19 & 8.05 & 4.76 & 2.94 & 29.45 \\
T5-WTA              & 18.53 & 9.98 & 6.06 & 3.92 & 32.96 \\
HTA-WTA             & 22.15 & \textbf{14}.\textbf{29} & \textbf{10}.\textbf{19} & \textbf{7}.\textbf{67} & \textbf{34}.\textbf{82} \\
\hline
\end{tabular}
\caption{\label{tab:results_main} Models' performances (percentages) on the collected dataset. For all scores, higher is better.}
\end{table*}

Regarding the generated questions type, in Table \ref{tab:results_question_type} we show the performance of the T5-based models per question type (inferential and literal). Though \textit{One-Step} and \textit{HTA-WTA} models were trained on the same amount of data, the results show that \textit{HTA-WTA} model clearly performs  better than the \textit{One-Step} model, especially on inferential questions. We see a similar scenario when comparing \textit{One-Step} and \textit{T5-WTA} models, yet, the gap is smaller. In general, we can notice that the performance gaps for the inferential questions are larger than the literal ones. Thus, we can conclude that \textit{HTA-WTA} is generating more correct inferential questions, which is challenging. This experiment concludes that transformers-based models are capable of asking questions beyond the literal meaning of the text. This confirms what was shown by~\citet{liu2021probing} regarding the skills that language models can acquire. Additionally, as some training questions directly quote text from the given story. The T5 model was able to learn how to quote the proper segment of the passage when generating questions.

\begin{table}%[h]
\small
\centering
\begin{tabular}{l|c|c}
\hline
\textbf{Model}       & \textbf{Inferential} & \textbf{Literal} \\ 
\hline
One-Step    & 28.44 & 30.63 \\
T5-WTA      & 33.13 & 32.78 \\
HTA-WTA     & 35.45 & 34.08 \\
\hline
\end{tabular}
\caption{\label{tab:results_question_type} T5-based models' performances (percentages) on each question type using BLEURT metric.}
\end{table}

The \textit{One-Step} model performs similarly to the baselines, although it has been trained using the T5 model and on all three datasets. This may be due to the fact that we did not include the skill name in the encoder, which guides the model to generate skill related questions. To better understand the differences between the outputs of \textit{One-Step} and \textit{HTA-WTA} models, we used human evaluation. This evaluation is to assess the quality of the generated question in terms of \textit{1. Answerability (Ay), 2. Fluency (Fy), and 3. Grammaticality (Gy)} categories, following~\citet{harrison2018neural,azevedo2020exploring}. We include these three criteria as questions may have high \textit{Fluency} and \textit{Grammaticality} scores, but not be answerable. We select a sample of 110 story-question pairs from the test dataset, for both models. Then, we perform a human evaluation using crowdworkers on Amazon Mechanical Turk. We use a "master" qualification criteria to restrict the participation of workers in our evaluation study to those who have a high historical HIT accuracy, and workers are required to be located in an English speaking country. Each HIT was answered by three workers. Each worker needs reads the story, and provides ratings (1-5, low to high) for the generated questions, and the three criteria. Table \ref{tab:human_eval_1} shows the average rating assigned by the workers for the 3 criteria. Originally, we hypothesized that adding the skill name to the input would force the model to formulate a specific SBRCS question, even if it is not applicable to the current passage. Omitting the skill name may allow the model score high values as it has been left to decide the question. The results show that both models are similar in terms of the given categories, except that \textit{HTA-WTA} performs slightly better in all of the three categories. However, these results refute our claim and show that adding the skill information makes the model generates slightly better questions in terms of quality. \bilal{In Section \ref{sec:appendix:ablation}, we present an ablation test and discuss some causes of errors in generating questions.}

\begin{table}%[h]
\small
\centering
\begin{tabular}{lcccc}
\hline 
\textbf{Model} & \textbf{Ay} & \textbf{Fy} & \textbf{Gy} & \textbf{Skills Accuracy} \\ \hline
One-Step    & 3.82 & 4.28 & 4.37 & 0.16 \\
HTA-WTA     & 3.89 & 4.29 & 4.45 & 0.8 \\
\hline
\end{tabular}
\caption{\label{tab:human_eval_1} Human evaluation ratings for our 3 criteria, on a scale 1-5.}
\end{table}

\noindent \textbf{Impact of Skill Name Token.} In order to quantify the impact of skill name in the input, we do another human manual evaluation to assess \bilal{how beneficial the skill name token is when we add it to the \textit{HTA-WTA} model}.
Thus, we ask two professional persons who were involved in the annotation process to assign skill names to the generated questions of both \textit{One-Step} and \textit{HTA-WTA} models. \bilal{We selected these models as they were trained on the same amount of data; the only difference between them is that the \textit{HTA-WTA} model uses the skill name token}. We utilize the same question sample that was used in the previous human evaluation experiment. Few annotation conflicts were found and were solved after a discussion. We evaluate the results using accuracy (see Table \ref{tab:human_eval_1}). The result for \textit{One-Step} model is 0.16, and 0.8 for \textit{HTA-WTA} model. We can clearly see a large gap in accuracy between both models, and this becomes clear with the skills that have a low number of instances in the dataset (e.g. Figurative Language, Predicting, etc.). \bilal{This result shows that, in addition to using the skill name token to control the skill of the generated questions, it helps the model to learn the underrepresented skills in the dataset}. Table \ref{tab:apndx:fine:grained} in Appendix \ref{sec:appendix:fine:grained} presents the F1 scores per skill name. We also notice that \textit{HTA-WTA} model performed perfectly on the given sample of \textit{Predicting} and \textit{Figurative Language} (F1 is 1.0 for each skill). This is an interesting result given that the type of the questions for both skills is inferential, which is harder to generate compared to the literal questions.

\begin{figure}%[H]
\centering
\includegraphics[width=8.5cm]{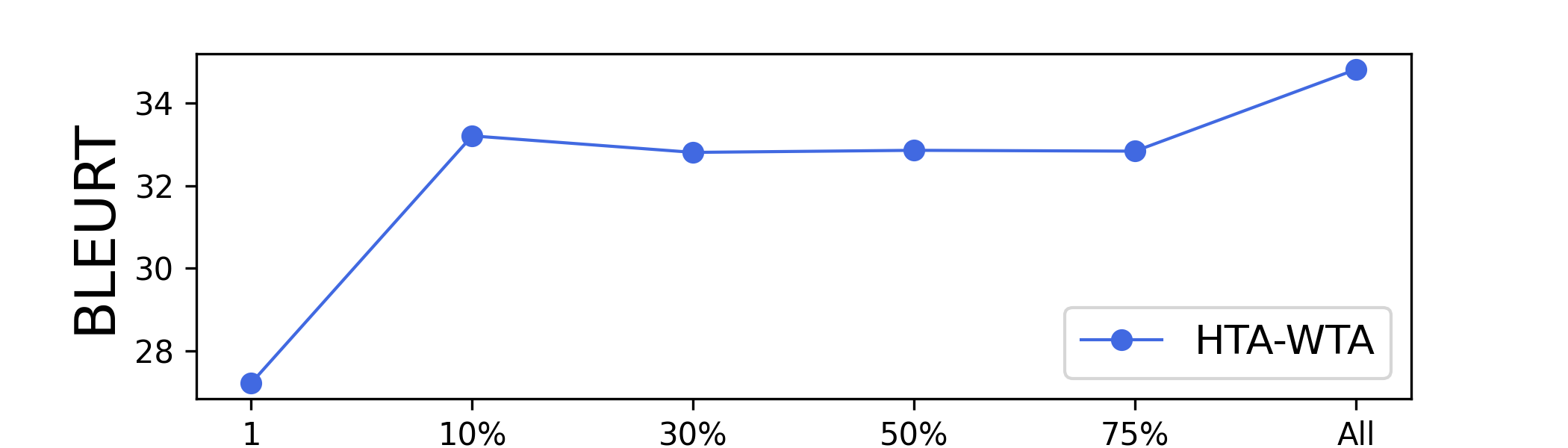}
\caption{Few-shot performance in BLEURT of the \textit{HTA-WTA} model over a percentage of added few-shot samples. 1 means single instance per skill (9 instances).}
\label{fig:few:shot}
\end{figure}

\noindent \textbf{Few-Shot Generation.} The process of manually writing questions to assess humans SBRCS is difficult. In some stories, professional writers find obstacles in writing questions for some skills as those skills require high attention and advanced reasoning skills to be written. We can see that in our own dataset, as some skills have fewer questions (e.g. Predicting, Visualizing, etc.). Thus, in this experiment, we evaluate the performance of \textit{HTA-WTA} model when we inject a low percentage of the skills' instances into the training set. This experiment will simulate the case when training a model on a dataset that contains few skills' instances. We use the stratified sampling technique when sampling fewer instances from the collected dataset. Figure \ref{fig:few:shot} shows that injecting only 10\% of the data led to a boost in performance of 5.99 (BLEURT). The result at 10\% (33.21\%) exceeds the results of most of the baselines and is higher than \textit{T5-WTA} and \textit{NQG-MAX} models when trained on all the datasets (see Table \ref{tab:results_main}). In Table \ref{sec:appendix:few:shot} in the appendix, we present the results considering other models and metrics. In most cases, the performance gradually improves as data grows. We notice a small drop when we move from 10\% to 30\%. This behaviour was previously reported by~\citet{stappen2020cross}. Further research is needed to investigate the causes of this behaviour.

% \vspace{0.5em}
\section{Conclusion and Future Work}
\label{sec:conclusion}
In this paper, we presented a new reading comprehension dataset to assess reading skills using stories. Unlike previous datasets that focused on either inferential or literal questions, our dataset has nine different SBRCS, each contains inferential and literal questions. In addition to that, we proposed \textit{HTA-WTA} model which uses two-steps fine-tuning processes to take advantage of previous datasets which have different question formats, and to learn how to ask skill-related questions. We evaluated the model on the collected dataset and compared it to several strong baselines. Our extensive experiments showed the effectiveness of the model. Additionally, \textit{HTA-WTA} is able to generate high quality questions when only 10\% of the dataset is used ($\sim$240 instances). In future work, we plan to extend our dataset with additional skills, and to investigate how our model can be integrated into online educational platforms.
%we can apply zero-shot generation to generate new kinds of skills.

% \section*{Acknowledgements}

\section{Ethical Considerations}
\label{ethical}

\noindent \textbf{Data collection and Annotation.} 
We made sure that the sources we use to collect stories do not prevent any kind of copyright infringement. The content distribution licenses were checked before any use. Additionally, we manually examined the stories and the created questions to ensure there are no privacy or ethical concerns, e.g., toxic language, hate speech, or any bias against underrepresented groups. EyeRead has outreach programs in place to recruit writers from diverse populations, incorporate their writing into the online system, and properly compensate them for their work. Writers that created questions earned comparable hourly wages to those earned by salaried teachers in a summer program. We estimated the amount of time AMT workers need to finish a HIT and then we compensated them so that the payment rate was higher than the local living wage per hour. Each AMT worker received \$0.41 USD for completing one HIT, which we estimated would take 1 minute.

\noindent \bilal{\textbf{Bias in Language Models.} Recently, many research works found that language models have several types of bias, e.g. gender, race, religion, etc., and this is due to the data used to train them \cite{liang2021towards}. Removing bias from language models completely is difficult, if not impossible \cite{gonen2019lipstick}. Thus, here we acknowledge that the QG model we trained might cause ethical concerns, e.g. generating biased questions about stories' characters. EyeRead is keenly aware of this, and continues to monitor both teacher and model-generated questions before they are integrated into their system.}

\bibliography{anthology}
\bibliographystyle{acl_natbib}

\appendix

\section{Appendix}
\label{appendix}

\subsection{Further Details on Skills}
\label{sec:appendix:skills:details}
\bilal{In the following, we elaborate more on the reading comprehension skills:} %with adding examples from the collected dataset:}

\begin{enumerate}
    \item \noindent \textbf{Basic Story Elements (BSE)}: 
    Determining what are the main story elements is one of the comprehension skills to assess the reader understanding. Using this skill, we can understand whether the reader is able to identify the main characters and environment settings of the stories.
    
    \item \noindent \textbf{Character Traits (CT)}: 
    Identifying permanent traits that can be assigned to characters or describe character development. For instance, knowing what most likely \textit{X} character felt during the story, recognizing facts about \textit{X}, identifying main adjectives that \textit{X} has, etc.
    
    \item \noindent \textbf{Close Reading (CR)}:
    Identifying the place in a story where the author best describes or explains a key point. Also, it includes questions to identify the purpose of a quote or a sentence. This skill requires advanced reading comprehension ability from the reader since its answers cannot be extracted directly from the story text, where inferential skills are needed.
    
    \item \noindent \textbf{Figurative Language (FL)}: 
    Figurative language is common in stories as it makes ideas and concepts easier to visualize by the reader. Also, it is an effective way of conveying an idea that is not easily understood. With this skill, we examine the reader ability of recognizing the implicated meaning of a sentence or a type of figurative language.
    
    \item \noindent \textbf{Inferring (I)}: 
    Writers sometimes jump into the action or skip forward in their stories. Good readers must infer what happened in between scenes if the time in-between is not explicitly detailed. In addition, readers must infer their characters' emotions if their characters do not share those aloud.
    
    \item \noindent \textbf{Predicting (P)}:
    Predicting involves guessing what will happen next. It is different from inferring; inferring is guessing what is happening now or what happened before. Good readers do not let books passively happen to them, they work to "solve" the story before it reaches its end by finding clues and using them to guess what will happen next or to guess how the conflict will be resolved.
    
    \item \noindent \textbf{Summarizing (S)}: Consolidating a text into a precise synopsis of only the most key information. Summarizing skill contains the main literary elements of the characters, the problem, and the solutions. Key events from the beginning, middle, and end are included in a summary.
    
    \item \noindent \textbf{Visualizing (V)}: This skill requires readers to visualize scenes in their heads to fully comprehend the story. It can assess readers ability of imagining specific events or elements in the stories.
    
    \item \noindent \textbf{Vocabulary (VO)}: 
    Identifying the meaning of unfamiliar words in the text is a key skill for readers to fully comprehend the story. In this skill, the reader should identify the right meaning of a word within a context when the word has multiple possible definitions. Additionally, the reader should be able to identify vocabulary based questions related to identifying synonyms, antonyms, homophones, compound words, and word types (e.g. noun, verb, etc.).
    
\end{enumerate}

\subsection{The Theory Behind Skills Taxonomy}
\label{sec:appendix:taxonomy}
\bilal{There are three major approaches within literacy education to which teachers or schools subscribe: the \textit{whole-language approach} \cite{froese1996whole} (which is the idea that if teachers simply give kids books, kids will learn how to read), the \textit{structural literacy approach} \cite{moats2019structured} (which is the theory that letters sounds, words parts, and grammar rules must all be explicitly taught in order for students to be able to read successfully), and the \textit{balanced literacy approach} \cite{asselin1999balanced} (which basically blends the aforementioned two theories together, in the sense that students read authentic literature while also receiving targeted instruction in skills or strategies). In this work, we chose to use the \textit{balanced literacy approach} as it benefits from both approaches and as it is the newest approach.}

\bilal{At the beginning, we reviewed some of the most commonly used balanced literacy curricula that were released by publishing houses and universities. In particular, we devoted a lot of focus to the Readers and Writers Workshop Model\footnote{\href{https://readingandwritingproject.org/}{https://readingandwritingproject.org}} which is developed at Columbia University Teachers College, and to the documentations about reading levels that developed by Scholastic publishing house\footnote{https://www.scholastic.com/teachers/teaching-tools/book-lists/guided-reading-levels-o-p-book-list.html}. The Readers and Writers Workshop curricula were highly instrumental to us in breaking reading comprehension into sub-skills. Also, it is one of the most commonly used and referenced curricula among teachers. We reviewed the workshop materials to create a list of all of the skills that the workshop program highlighted. Then, we matched those against what was offered by Scholastic. This helped us create our primary list of skills. In this study, we are experimenting with nine skills out of around twenty skills. In this phase of the study, we are focusing on the most comprehensive and common skills. In the future, we will expand our work to include the rest of the skills.}

\subsection{Additional Data}
\label{sec:appendix:additional_data}
In addition to the collected dataset, we use two well-known datasets, SQuAD and CosmosQA. We choose these two datasets because of their large size, and their focus on literal or inferential questions.

\noindent \textbf{SQuAD} A reading comprehension dataset, consists of questions created by crowdworkers on a set of Wikipedia articles that cover a large set of topics (from musical celebrities to abstract concepts), where the answer to every question is a span from the corresponding reading passage~\cite{rajpurkar2016squad}. This dataset can be considered as an extractive QA dataset. It is one of the largest QA datasets in the literature. In this work, we use SQuAD 2.0 version with discarding the questions that have no answers. The size of the dataset is ~100K paragraph/question/answer triplets.

\noindent \textbf{CosmosQA} It is another reading comprehension dataset consisting of 35.6K paragraph/question pairs that require commonsense-based reading comprehension. It is a collection of people's everyday narratives, and it asks questions about the likely causes of events that require reasoning~\cite{huang2019cosmos}. We discard questions that have no answers in this dataset, resulting in ~28K paragraph/question/answer triplets.

\subsection{Ablation Test and Error Analysis}
\label{sec:appendix:ablation}
\bilal{\noindent \textbf{Ablation Test.} The results of our experiments confirmed the importance of both the skill name token and the two-steps training method. To quantify the impact of including the skill name token, we run \textit{T5-WTA} without including the skill name token (\textit{T5-WTA-unskilled}). 
We compare the \textit{T5-WTA-unskilled} to the \textit{One-Step} model; the only difference between these models is that \textit{One-Step} model includes SQuAD and CosmosQA datasets in the training data. The ablation test results in Table \ref{tab:ablation} shows that the skill name token and the additional training data both increase model performance.  \textit{T5-WTA-unskilled} BLEURT performance is lower than the BLEURT scores of the other two models.}

% I reworded this a fair bit above, so left the old wording here
%We compare the \textit{T5-WTA-unskilled} to the \textit{One-Step} model to study the impact of including the additional training data in the \textit{One-Step} model; both models are the same, and the only difference is that \textit{One-Step} model includes the additional two datasets in the training data. Table \ref{tab:ablation} shows the results of the ablation test. Briefly, we can see that both, the skill name token and the additional training data, are beneficial and improve the models' performances. We can notice that the \textit{T5-WTA-unskilled} model's result, considering BLEURT metric as an example, is lower than the other BLEURT scores of the two models.}

\begin{table*}%[h]
\small
\centering
\begin{tabular}{l|c|c|c|c|c}
\hline
\textbf{Model} & \multicolumn{1}{|c|}{\textbf{BLEU-1}} & \multicolumn{1}{|c|}{\textbf{BLEU-2}} & \multicolumn{1}{|c|}{\textbf{BLEU-3}} & \multicolumn{1}{|c|}{\textbf{BLEU-4}} & \multicolumn{1}{|c|}{\textbf{BLEURT}} \\ \hline
One-Step            & 15.19 & 8.05 & 4.76 & 2.94 & 29.45 \\
T5-WTA              & 18.53 & 9.98 & 6.06 & 3.92 & 32.96 \\
T5-WTA-unskilled    & 14.65 & 8.31 & 4.37 & 2.39 & 29.02 \\
\hline
\end{tabular}
\caption{\label{tab:ablation} The ablation test results (percentages).}
\end{table*}

\bilal{ \noindent \textbf{Error Analysis.} Here we are interested in further understanding the \textit{HTA-WTA} model's performance. We manually examined several generated questions to understand the sources of its errors. Given the unbalanced status of the dataset, we found that the model does not always generate an appropriate question for a given skill name, especially when that skill is underrepresented in the data (e.g. Visualizing, Figurative Language, etc.). In some cases, the model learned the style of the skill's questions, but in the given context, the generated question could not be answered. As an example, the following generated figurative language question quoted a sentence from a story about the space. The sentence is an event in the story and not a figurative language:}

\begin{displayquote}
    \centering
    \textit{\bilal{Which figurative language technique is being used in the phrase ``The first safe trip into space''?}}
\end{displayquote}

\bilal{This happens even for very common skill categories, again due to the difficulty (or even impossibility) of generating questions for some skill and story pairs. The other kind of error is the subjectivity in selecting the "correct" words from the story. For instance, giving the following \textit{Vocabulary} question from the dataset:}

%again, I made big changes above, so I leave the original here for your reference
%Language models proved to be a very good language generator as they have been evaluated on several text generation tasks. Nonetheless, the models' performances contrast from one task to another due to different reasons. To understand \textit{HTA-WTA} model's performance more, we manually examined its generated questions to understand the causes of errors. Given the unbalanced status of the dataset, we found that the model does not always generate the correct question for a given skill name that is underrepresented in the data (e.g. Visualizing). We noticed in some cases that the model learned only the style of the skill's questions, but considering the context, the question cannot be answered. Actually, we noticed that this kind of errors happen sometimes even with the skill categories that are majority in the dataset. We argue that this is due to the difficulty of generating a question for that given skill and story. The other kind of error is the subjectivity in selecting the "correct" words from the story. For instance, giving the following \textit{Vocabulary} question from the dataset:

\begin{displayquote}
    \centering
    \textit{\bilal{What is the correct definition of the word "decoy" as it is used in the story?}}
\end{displayquote}

\bilal{For this kind of question, annotators chose words that can have multiple meanings, some of which may be unfamiliar to school children. The process of choosing those words is subjective. Although both annotators agreed on the word in the previous example, the model chose to select another word from the story ("panting"). In other cases, the question asks about the definition of a word within a sentence from the story (e.g. What is the meaning of ``word'' as it is used in this sentence: ``quoted sentence'').  We noted that when the model generated the question, it selects the correct word but sometimes used a randomly quoted sentence from the story that didn't contain the word.}

\subsection{Manual Evaluation Results of Questions' Skills}
\label{sec:appendix:fine:grained}
In Table \ref{tab:apndx:fine:grained}, we show the fined-grained results per skill name after the manual labeling experiment for the generated questions from both \textit{One-Step} and \textit{HTA-WTA} models.

\begin{table*}%[h]
\small
\centering
\begin{tabular}{l|c|c|c|c|c|c|c|c|c}
\hline
& \textbf{BSE} & \textbf{CT} & \textbf{CR} & \textbf{FL} & \textbf{I} & \textbf{P} & \textbf{S} & \textbf{V} & \textbf{VO} \\
% & \multicolumn{1}{c|}{Basic} & \multicolumn{1}{c|}{\multirow{2}{*}{Char-}} & \multicolumn{1}{c|}{\multirow{2}{*}{Close}} & \multicolumn{1}{c|}{\multirow{2}{*}{Figura-}} & \multicolumn{1}{c|}{\multirow{3}{*}{Infer-}} & \multicolumn{1}{c|}{\multirow{3}{*}{Predi-}} & \multicolumn{1}{c|}{\multirow{3}{*}{Summ-}} & \multicolumn{1}{c|}{\multirow{3}{*}{Visua-}} & \multicolumn{1}{c}{\multirow{3}{*}{Vocab-}} \\
% Model & \multicolumn{1}{c|}{Story} & \multicolumn{1}{c|}{\multirow{2}{*}{Traits}} & \multicolumn{1}{c|}{\multirow{2}{*}{Read-}} & \multicolumn{1}{c|}{\multirow{2}{*}{Lang-}}  &  &  &  &  &  \\
% & \multicolumn{1}{c|}{Elem-} &  &  &  &  &  &  &  &  \\
\hline
\#instances & 12 & 8 & 23 & 7 & 14 & 6 & 14 & 10 & 16 \\ \hline
One-Step    & 0.13 & 0.00 & 0.31 & 0.00 & 0.19 & 0.00 & 0.07 & 0.00 & 0.18 \\
HTA-WTA     & 0.88 & 0.93 & 0.68 & 1.00 & 0.69 & 1.00 & 0.81 & 0.18 & 1.00 \\
\hline
\end{tabular}
\caption{\label{tab:apndx:fine:grained} F1 score results per skill name.}
\end{table*}

\subsection{Few-Shot Question Generation Results}
\label{sec:appendix:few:shot}
In Table \ref{tab:apndx:few_shot}, we show the few-shot experiment's results considering both scoring metrics (BLEU, and BLUERT). We do not experiment with \textit{One-Step} model as we need to sample SQuAD and CosmosQA datasets when we sample the collected data; it is hard to set up a fair comparison here as, for instance, sampling 10\% of SQuAD dataset is larger than the whole collected dataset.

\begin{table*}%[h]
\small
\centering
\begin{tabular}{l|c|c|c|c|c|c}
Instances Ratio & Model & BLEU-1 & BLEU-2 & BLEU-3 & BLEU-4 & BLEURT \\ \hline
1       & T5-WTA  &  8.61  &  3.38  &  1.71  &  1.04  &  24.47  \\
1       & HTA-WTA &  10.2  &  4.74  &  2.85  &  1.96  &  27.22  \\
\hline
0.1     & T5-WTA  &  14.8  &  6.68  &  3.63  &  2.22  &  29.09  \\
0.1     & HTA-WTA &  16.55 &  9.54  &  6.28  &  4.37  &  33.21  \\
\hline
0.3     & T5-WTA  &  16.02 &  8.3   &  5.07  &  3.45  &  29.69  \\
0.3     & HTA-WTA &  16.14 &  9.7   &  6.64  &  4.82  &  32.81  \\
\hline
0.5     & T5-WTA  &  16.32 &  8.25  &  4.77  &  3.00  &  31.20  \\
0.5     & HTA-WTA &  15.48 &  9.25  &  6.34  &  4.61  &  32.86  \\
\hline
0.75    & T5-WTA  &  18.9  &  10.12 &  6.24  &  4.19  &  32.65  \\
0.75    & HTA-WTA &  18.69 &  11.53 &  7.97  &  5.74  &  32.84  \\
\hline
All     & T5-WTA  &  18.53 &  9.99  &  6.07  &  3.93  &  32.96  \\ 
All     & HTA-WTA &  22.15 &  14.3  &  10.2  &  7.67  &  34.82  \\ 
\hline
\end{tabular}
\caption{\label{tab:apndx:few_shot} Few-shot performance (percentages) of the \textit{HTA-WTA} and \textit{T5-WTA} models over a percentage of added few-shot samples. 1 means single instance per skill (9 instances).}
\end{table*}

\subsection{Samples of the Generated Questions}
\label{sec:appendix:sample}
In this section, we list some random examples from \textit{HTA-WTA} model for inferential questions:

\begin{itemize}
    \item Story: \textquote{The Line 1 Toronto train was a subway like many others you’ve seen. He rocketed down Yonge Street, around the Union loop, and rattled off towards Vaughn. At Vaughn he'd let out a loud, hissing sigh and a clanking sort of grunt, then reverse and do the whole thing backwards all over again. He liked his transit union job well enough, but he couldn’t help thinking about the lights at the end of his tunnels. No matter how long he'd been running, or how much he wished for anything else, that little hopeful point of light always turned out to be just one more dirty subway platform.}
    
    Generated \textit{Figurative Language} question: \textquote{Reread this sentence: "He rocketed down Yonge Street, around the Union loop, and rattled off towards Vaughn." Which figurative language technique is being used here?}
    
    Generated answer: \textquote{Alliteration}.

    \item Story: \textquote{"The map says left", said Bri.  "But my heart says right!" cried Rob.  "Is your heart full of hidden treasure?" asked Bri. "Yes." Rob replied. "At least, that's what my mom says."}
    
    Generated \textit{Inferring} question: \textquote{Why do you think Bri's heart says "But my heart says right!"?}
    
    Generated answer: \textquote{Because she thinks she has found something}.

    \item Story: \textquote{Mary looked at it, not really knowing why the hole was there, and as she looked she saw something almost buried in the newly-turned soil. It was something like a ring of rusty iron or brass and when the robin flew up into a tree nearby she put out her hand and picked the ring up. It was more than a ring, however; it was an old key which looked as if it had been buried a long time.  Mistress Mary stood up and looked at it with an almost frightened face as it hung from her finger.  "Perhaps it has been buried for ten years," she said in a whisper. "Perhaps it is the key to the garden!"}
    
    Generated \textit{Vocabulary} question: \textquote{Reread this sentence: "Perhaps it has been buried for ten years" What is the correct definition of the word "frightened" as it is used here?}
    
    Generated answer: \textquote{Scared}.
\end{itemize}

\end{document}